\def\year{2019}\relax
\documentclass[letterpaper]{article} %DO NOT CHANGE THIS
\usepackage{aaai19}  %Required
\usepackage{times}  %Required
\usepackage{helvet}  %Required
\usepackage{courier}  %Required
\usepackage{url}  %Required
\usepackage{graphicx}  %Required

\usepackage{multirow}

\usepackage{algorithm}
\usepackage{algorithmic}
\newcommand{\fracpartial}[2]{\frac{\partial #1}{\partial #2}}

\begin{document}
\title{SepNE: Bringing Separability to Network Embedding \thanks{This work was supported by the National Natural Science Foundation of China (Grant No. 61876006 and No. 61572041).}}
\author{
Ziyao Li,\textsuperscript{\rm 1}
Liang Zhang,\textsuperscript{\rm 1}
Guojie Song\textsuperscript{\rm 2}\thanks{Corresponding author. Email: gjsong@pku.edu.cn} \\
\textsuperscript{\rm 1}Yuanpei College, Peking University, China \\
\textsuperscript{\rm 2}Key Laboratory of Machine Perception, Ministry of Education, Peking University, China\\
\{leeeezy, zl515, gjsong\}@pku.edu.cn
}

\maketitle
\frenchspacing
\begin{abstract}
Many successful methods have been proposed for learning low dimensional representations on large-scale networks, while almost all existing methods are designed in inseparable processes, learning embeddings for entire networks even when only a small proportion of nodes are of interest. This leads to great inconvenience, especially on super-large or dynamic networks, where these methods become almost impossible to implement. In this paper, we formalize the problem of separated matrix factorization, based on which we elaborate a novel objective function that preserves both local and global information. We further propose SepNE, a simple and flexible network embedding  algorithm which independently learns representations for different subsets of nodes in separated processes. By implementing separability, our algorithm reduces the redundant efforts to embed irrelevant nodes, yielding scalability to super-large networks, automatic implementation in distributed learning and further adaptations. We demonstrate the effectiveness of this approach on several real-world networks with different scales and subjects. With comparable accuracy, our approach significantly outperforms state-of-the-art baselines in running times on large networks.%Our approach significantly outperforms state-of-the-art baselines in both speed and performance.
\end{abstract}

\section{Introduction}
Learning low dimensional representations of network data, or network embedding (NE), is a challenging task on large networks, of which the scales can reach billion-level and is growing rapidly. For example, the number of monthly-active users of Facebook reaches $2.23$ billion and increases $11\%$ yearly. \footnote{\url{https://investor.fb.com/investor-news/}}
At the same time, although sizes of networks may infinitely grow as data accumulates, it is often the case that only small proportions of nodes are of interest in downstream applications. This is the starting point of this paper: can we respectively learn representations for different subsets of nodes-very small compared to the collectivity-while preserving information of the entire network? If so, we can obtain good representations for the requested nodes without the redundant efforts to embed irrelevant ones.

Efficiency is a major aspect of contemporary NE studies, and various methods that are applicable to large-scale networks have been proposed \cite{deepwalk,line,node2vec}. Almost all of these methods embed entire networks with inseparable processes, in which the representation of one node depends on represented outcomes of every other node. A globally-defined optimum can be achieved under this framework, while it also causes great inconvenience: the maximum network size such methods can handle is eventually limited. For example, it takes LINE~\shortcite{line}, one of the fastest algorithms, several hours to embed a million-level network. Thousands of hours may be spent to achieve equivalent performance on billion-level networks. Another type of methods learns models in an inductive manner and conduct inferences over unseen data \cite{graphsage}. These models have convenient inference processes, while they cannot variate over time and rely on large training data and time to achieve good performance.

\begin{figure}
  \centering
  \includegraphics[width=0.4\textwidth]{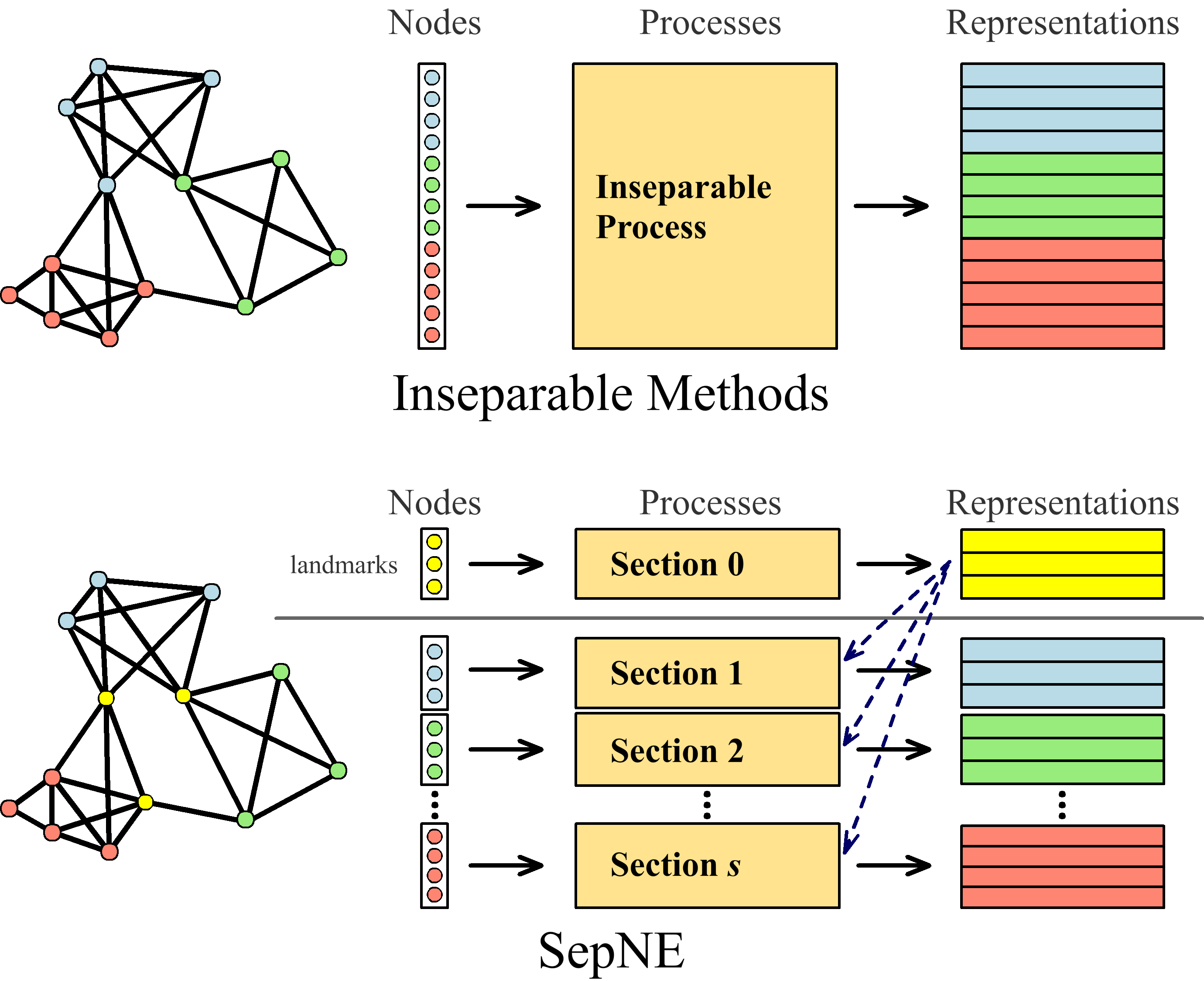}
  \caption{Inseparable and separable NE processes.}\label{fig:illustrate}
\end{figure}

The efficiency problem over super-scale networks is impossible to solve directly, as the running time of algorithms inevitably grows proportionally to problem scales. Therefore, we bring up a new perspective of solving efficiency problems: \textit{separability}.
The separability of an algorithm indicates an ability of being conducted by different workers without exchanging information and merging outputs. In plain words, a separable algorithm divides the original problem into different \textit{self-standing} sub-problems and separately solves each, and the solution to the sub-problems are directly usable answers instead of intermediate results. As networks are naturally composed of nodes and their relationships, an instinctive way to design separable NE algorithms is to partition the entire node set into small subsets and to separably embed each set. The solutions to the sub-problems yield direct meanings as the representations of the corresponding set of nodes.

In this paper, we implement separability in NE problems under matrix-factorization-based framework. We formalize the problem of separated matrix factorization (SMF) and elaborate a novel loss function that preserves matrix information on local, global and landmark level. We then propose \textbf{SepNE} (SEParated Network Embedding), a separable NE method based on SMF. Figure~\ref{fig:illustrate} illustrates the major difference between SepNE and existing methods. SepNE first partitions the node set into small subsets. A special \textit{landmark} set is then established as references for partitioned subsets to implement \textit{comparability}, that representations of different sets lie in the same linear space. After the landmarks are embedded, representations of different subsets are derived from the objective function defined in SMF.

Separability in NE problems yields several specific advantages. Firstly, separability makes it available to embed only the requested nodes and thus reduces the vain efforts in embedding irrelevant ones; in addition, the optimization complexity of SepNE is relevant only to the number of requested nodes instead of the entire network scale, leading to scalability to super-large networks. Secondly, even if entire networks are on request, SepNE shows higher speed than state-of-the-art algorithms due to its simplicity, while yielding comparable accuracy. Thirdly, separability leads to automatic implementations in multi-agent systems and further adaptations in dynamic networks. We evaluated SepNE with real-world networks of different sizes and topics. With equivalent or better accuracy, SepNE is at least 50\% faster than current NE methods including DeepWalk~\shortcite{deepwalk} and LINE.

A potentially more important contribution of this paper is the generalization of SMF. Maintaining competent performance, SMF reduces the complexity of MF problems from cubic to almost linear. This leads to intriguing further applications in the massive collectivity of MF-based algorithms.

%The remainder of the paper is organized as follows. We formalize SMF in Section 2. The detailed model is presented in Section 3, and the experiment outcomes are analyzed in Section 4. Related work is presented in Section 5. In Section 6, we summarize our work and propose future work.

\section{Separated Matrix Factorization}
\subsection{Preliminaries}
Given a matrix $M$, matrix factorization (MF) aims to find two matrices $W$ and $C$ that both satisfy given constraints and minimize the residuals of reconstructing $M$ with $\tilde{M}=W^TC$. Denoted in formulas, we have
\begin{equation}
  \min_{W, C}
  \Vert M - W^TC \Vert. \label{eqn:matfac}
\end{equation}
$W,C$ have lower ranks than $M$. In the embedding task of an $n$-node network, $M$ is of size $(n \times n)$, in which each entry indicates a \textit{proximity} between the two corresponding nodes. The proximities can be defined in various metrics, such as edge weights between nodes. Columns in $W$ are desired representations. Columns in $C$ are interpreted as representations of nodes when regarded as \textit{contexts}.

\subsection{Problem Definition}
Directly factorizing matrices of large-scale networks can be unacceptably time-costly. Therefore, we propose SMF, a new optimization problem, as a trade-off between speed and accuracy. To implement separability, SMF divides the problem with a partition over the node set, and correspondingly partitions the proximity matrix. Below is a formal definition of separated matrix factorization (SMF) in NE scenarios.

SMF takes a network $G=(V,E),|V|=n$, its proximity matrix $M$, and a partition setup $f:V \mapsto \mathcal{V}$, $\mathcal{V}=\{V_1, \cdots , V_s\}$ as inputs. The task is to derive representations $(W_1, \cdots, W_s)$ and $(C_1, \cdots, C_s)$ for the partitioned sets that optimally reconstruct $M$. The loss of the reconstruction is defined the same as Problem (\ref{eqn:matfac}). \footnote{Frobenius forms are usually adopted in SMF, since the Frobenius form of a matrix is additive of all its entries, and therefore can be decomposed into sums of all the Frobenius forms of its partitioned sub-matrices.} Without loss of generosity, we permute and partition $M$ according to $\mathcal{V}$ as
\begin{displaymath}
  M = \left(
  \begin{array}{ccc}
    M_{11} & \cdots & M_{1s} \\
    \vdots & \ddots & \vdots \\
    M_{s1} & \cdots & M_{ss}
  \end{array}
  \right),
\end{displaymath}
where $M_{ij}$ indicates the proximities between $V_i$ and $V_j$. To achieve independence between sections, SMF restricts that the embedding section of every set is conducted (i) with only the proximities related to itself (the section of embedding $V_i$ can leverage only $M_{ij}$ and $M_{ji}$, $j=1,\cdots,s$), and (ii) without any outcomes of other sections.

\subsection{Method}
Partitioning the nodes is the first step to separability. However, representations in the partitioned sets can be incomparable due to the limitation over the access to proximities. In another word, representations of different sets do not have a unified constraint that bounds them in the same linear space. To implement comparability, we establish landmarks with highly interactive nodes in the network, which serve as invariant references for different subsets. For the factorization process, preserving only local information is a simple way to reconstructs micro-structures of networks at a loss of global references. Combining landmark solves the comparability problem, while it still ignores the interactions between different subsets. Therefore, we elaborate a novel objective function for SMF that preserves local, global and landmark information, which achieves state-of-the-art performances.

\subsubsection{Local information.} The proximities in the partitioned matrix are naturally divided into two types, namely \textit{local information} and \textit{global information}. Local information refers to the proximities within every set, or sub-matrices on the diagonal; global information refers to the proximities between all pairs of different sets, or the off-diagonal sub-matrices. We start modeling SMF with a na\"ive simplification that preserves only local information by factorizing $s$ matrices on the diagonal:
\begin{displaymath}
  \min_{W_i,C_i} \Vert M_{ii} - W_i^TC_i \Vert, i=1,\cdots,s.
\end{displaymath}
This primitive approach discards all interactions between different sets, leading to incomparable representations.

\subsubsection{Landmark information.} To implement comparability, we resort to a third type of information, \textit{landmark information}. Landmark information indicates the proximities between subsets and manually established \textit{landmarks}, a special set of nodes (denoted as $V_0$) that are chosen as references for different subsets. The improved approach sets a unified constraint over landmark information in different sets and solves the problem in two stages, formulated as
\begin{equation}
  \min_{\Phi,\Psi} \Vert M_{00} - \Phi^T\Psi \Vert, \label{eqn:stage1}
\end{equation}
\begin{eqnarray}
  \nonumber \min_{W_i,C_i}
  \left\Vert \left(
  \begin{array}{cc}
    M_{00} & M_{0i} \\
    M_{i0} & M_{ii}
  \end{array}
  \right) - \left(
  \begin{array}{cc}
    \Phi^T\Psi & \Phi^TC_i \\
    W_i^T\Psi & W_i^TC_i
  \end{array}
  \right) \right\Vert, &&\\
  i=1,\cdots,s. \label{eqn:stage2}&&
\end{eqnarray}
The first stage embeds the landmarks ($W_0=\Phi$, $C_0=\Psi$) in Problem (\ref{eqn:stage1}), and the second stage derives representations for rest sets by solving Problem (\ref{eqn:stage2}) with calculated $\Phi$ and $\Psi$. If Frobenius forms are used, the loss in Problem (\ref{eqn:stage2}) can be explicitly decomposed into local and landmark loss as
\begin{eqnarray}
  \mathcal{L}_i^{lc}(W,C) &=&
    \frac{1}{2} \Vert M_{ii} - W^TC \Vert_F^2, \label{eqn:lc} \\
  \nonumber \mathcal{L}_i^{lm}(W,C) &=&
    \frac{1}{2} \Vert M_{0i} - \Phi^TC \Vert_F^2 \\
     & & \quad + \frac{1}{2} \Vert M_{i0} - W^T\Psi \Vert_F^2. \label{eqn:lm}
\end{eqnarray}

\subsubsection{Global information.} To further combine global information into the objective function, we elaborate a global loss by first transforming Problem (\ref{eqn:stage2}) into an equivalent form. We denote $k:=|V_0|$ and assume calculated $\Phi, \Psi \in \mathbf{R}^{(d \times k)}$ are of rank $d$. \footnote{This can be guaranteed with SVD decomposition if $k \ge d$.} $W_i, C_i \in \mathbf{R}^{(d \times |V_i|)}$ can then be represented as linear combinations of columns in $\Phi,\Psi$, formulated in matrix denotation as
\begin{eqnarray*}
  W_i &=& \Phi A_i \\
  C_i &=& \Psi B_i, \quad i=1,2,\cdots,s,
\end{eqnarray*}
where $A_i, B_i \in \mathbf{R}^{(k \times |V_i|)}$ are the coefficient matrices.

Consider a simple case where $s=2$, $V_0$ is the set of landmarks and $V_1$ is the target subset to embed. After the transformation, global information is preserved through
\begin{eqnarray}
  && \min_{A_1,B_2} \Vert M_{12} - A_1^T\Phi^T\Psi B_2 \Vert \label{eqn:glb1} \\
  && \min_{A_2,B_1} \Vert M_{21} - A_2^T\Phi^T\Psi B_1 \Vert. \label{eqn:glb2}
\end{eqnarray}
Problem (\ref{eqn:glb1})(\ref{eqn:glb2}) are not separable, for the results of embedding $V_2$ ($A_2$,$B_2$) exist in $V_1$ problem. However, a surprising property emerges after the transformation, that $\Phi^T\Psi B_2=W_0^TC_2=\tilde{M}_{02}$ can be well-approximated with $M_{02}$ if representations of $V_2$ are required to preserve landmark information, $A_2^T\Phi^T\Psi$ similarly. Therefore, Problem (\ref{eqn:glb1})(\ref{eqn:glb2}) can be substituted as
\begin{eqnarray}
  && \min_{A_1} \Vert M_{12} - A_1^TM_{02} \Vert \label{eqn:subglb1} \\
  && \min_{B_1} \Vert M_{21} - M_{20}B_1 \Vert, \label{eqn:subglb2}
\end{eqnarray}
and separability is achieved.

The idea can be generalized to any given $s$ and $V_i$ by simply substituting all $V_2$-related variables to $V_{\bar{i}}$-related ones, where $V_{\bar{i}}=\bigcup_{j \notin \{0,i\}} V_j$. The approximation still holds if landmark information is preserved in all sets. For any set $V_i$, the global loss function is defined as
\begin{displaymath}
  \mathcal{L}_i^{gb} (A,B) = \frac{1}{2}(
  \Vert M_{i\bar{i}} - A^TM_{0\bar{i}} \Vert_F^2 +
  \Vert M_{\bar{i}i} - M_{\bar{i}0}B \Vert_F^2).
\end{displaymath}

\subsection{Final Optimization Problem}
Combined with $\lambda$-scaled global loss and regularization over $A,B$, the final loss function of SMF becomes
\begin{eqnarray}
  \mathcal{L}_i (A,B) &=& \mathcal{L}_i^{lc} (A,B) +
          \mathcal{L}_i^{lm} (A,B)\\
  \nonumber & & \quad + \lambda \mathcal{L}_i^{gb} (A,B)
                + \frac{\eta}{2} (\Vert A \Vert_F^2 +
                                  \Vert B \Vert_F^2),
  \label{eqn:total}
\end{eqnarray}
where $\mathcal{L}_i^{lc}$ and $\mathcal{L}_i^{lm}$ are redefined in $A,B$-denotation, namely
\begin{eqnarray}
  \mathcal{L}_i^{lc}(A,B) &=&
    \frac{1}{2} \Vert M_{ii} - A^T\Phi^T\Psi B \Vert_F^2, \\
  \nonumber \mathcal{L}_i^{lm}(A,B) &=&
    \frac{1}{2} \Vert M_{0i} - \Phi^T\Psi B \Vert_F^2 \\
     & & \quad + \frac{1}{2} \Vert M_{i0} - A^T\Phi^T\Psi \Vert_F^2.
\end{eqnarray}
Accordingly, the final optimization problem of SMF is formulate as
\begin{displaymath}
\begin{array}{ll}
  \displaystyle W_0=\Phi, & C_0=\Psi. \\
  \displaystyle W_i=\Phi A_i, & C_i=\Psi B_i, \\
  \multicolumn{2}{c}{\displaystyle A_i,B_i=\arg\min_{A,B} \mathcal{L}_i(A,B), \quad i = 1,2,\cdots,s.}
\end{array}
\end{displaymath}

\section{SepNE: Separated Network Embedding}
In this section, we propose SepNE, a simple and separable NE approach based on SMF. A general framework of SepNE is presented in Algorithm~\ref{alg:Framework}. We then illustrate the details of SepNE, including the partition setups, landmark-selecting approaches and optimization method.

SepNE takes a given network as input and outputs node representations. In the preparation stage, landmarks are selected and embedded, and rest nodes are partitioned under a certain setup. In the second stage, partitioned sets are independently embedded by optimizing the SMF problem. The second stage is designed in a separable manner, so that if a small proportion of nodes are requested, Loop \ref{step:loop} in Algorithm \ref{alg:Framework} can be conducted only on the sets containing these nodes. Besides, separability allows cycles in the loop to be run distributedly.

\begin{algorithm}[htb]
\caption{ General framework of SepNE.}
\label{alg:Framework}
\begin{algorithmic}[1]
\REQUIRE $G=(V,E), |V|=n$.
\ENSURE Node embeddings for partitioned sets of nodes.
\STATE Partition rest nodes in set $V$ into $s$ subsets as $\mathcal{V}$; \label{step:partition}
\STATE Sample $k$ landmarks as set $V_0$;
\label{step:sample}
\STATE Conduct SVD on calculated proximity matrix $M_{00}$ and calculate $\Phi,\Psi$: \\
\begin{center}
  $M_{00}=U_d\Sigma_d V_d^T$, \\
  $\Phi=U_d\sqrt{\Sigma_d}, \Psi=V_d\sqrt{\Sigma_d}$;
\end{center}
\label{step:svd}
\STATE $W_0=\Phi$;
\label{step:loop}
\FOR {$i=1,2,\cdots,s$}
  \label{step:calcmat}
  \STATE Calculate relevant proximity matrices $M_{0i}$, $M_{i0}$, $M_{ii}$, $M_{i\bar{i}}$, $M_{\bar{i}i}$;
  \label{step:optimize}
  \STATE Optimize the loss functions: \\
    \begin{center}
      $\displaystyle A_i,B_i=\arg\min_{A,B} \mathcal{L}_i(A,B)$;
    \end{center}
  \label{step:deriveEmbed}
  \STATE Calculate embeddings for set $V_i$: \\
    \begin{center}
      $W_i=\Phi A_i, C_i=\Psi B_i$
    \end{center}
\ENDFOR
\RETURN $(W_0,W_1,\cdots,W_s)$
\end{algorithmic}
\end{algorithm}

For the proximity matrix to be factorized, two metrics are adopted in SepNE. The first metric defines $M=A+A^2$, where $A$ is the transition matrix of PageRank~\shortcite{pagerank} \footnote{$A_{ij}=1/d_i$ if $(i,j)\in E$ and $0$ otherwise, where $d_i$ is the degree of node $i$.}. The second metric simplifies the first one with $M=I+A$. Our metrics are similar to TADW~\shortcite{tadw}, which proved an equivalency between factorizing $A$ and $A+A^2$ and DeepWalk~\shortcite{deepwalk} with very short walks. A more instinctive understanding of the metrics can be derived from the perspective of \textit{proximity orders}. $A$ can be interpreted as a measurement of first-order proximity, and $A+A^2$ a combination of first-order and second-order proximity. These concepts were proposed and further discussed in \cite{line}.

\subsection{Partition Setups}
We propose three different partition setups for SepNE in this paper. \textit{SepNE-LP} (Louvain Partition) partitions a network according to its communities using Louvain~\shortcite{louvain}. Leveraging community structures conforms \textit{matrix local information} to \textit{network local information} \footnote{Which refers to connections within real-world communities.}, which serves as an empirical approach to improve performance.

However, as SepNE leverages all the information of the proximity matrix, community-based partitions are not necessary. We further propose \textit{SepNE-RP} (Random Patition) which randomly assigns nodes to sets, and \textit{SepNE-IO} (Interested Only) which simply puts the requested nodes into one or more sets and ignores all unrequested ones.

\subsection{Landmark Selection}
Landmark-selecting approaches influence not only representations of the landmarks, but also the loss of all sets in the entire SMF problem. As the key intention of setting landmarks is to establish references for different sets, landmarks are expected (i) to have as much connection with rest nodes as possible; (ii) to have the connection cover as many sets as possible.

Approaches that select nodes with high degrees generally work well if $k$ is loosely controlled. However, on real-world networks, nodes with the highest degrees tend to distribute in a few giant and highly connected communities. When $k$ is strictly confined, choosing these nodes actually limits the number of sets these landmarks adjoin. To relieve this problem, we propose \textit{GDS} (Greedy Dominating Set), an approach that greedily maximizes the number of nodes the landmarks adjoins.

GDS first forms a maximum heap using degrees of nodes and initialize the landmark set as empty. After initialization, GDS iteratively examines the top of the heap. The top is simply removed if dominated by the current landmark set, otherwise added into the set and then removed. The process continues until the heap is empty or the size reaches $k$. Experiments show that GDS well captures the informative structure of a network when $k$ is strictly confined.

While serving as good references, landmarks selected with GDS are completely one-hop isolated. As a consequence, if only one-hop proximity is leveraged, $M_{00}$ of GDS is supposed to be a diagonal matrix. Furthermore, if $k>d$, SVD will generate null representations for some landmarks. Therefore, we only use GDS when higher order proximity is adopted or $k=d$. Otherwise, we implement degree-based approaches.

\subsection{Optimization}
The optimization problem in SepNE is solved similarly to \cite{yu14}, where $A,B$ are iteratively optimized as
\begin{eqnarray*}
  A^{(t+1)} &=& \arg \min_{A} \mathcal{L}_i(A, B^{(t)}), \\
  B^{(t+1)} &=& \arg \min_{B} \mathcal{L}_i(A^{(t+1)}, B).
\end{eqnarray*}
As explicit calculation of the loss function value involves large matrix multiplications, $A,B$ are calculated by solving the gradient-minimization problem $\fracpartial{\mathcal{L}}{A} = \fracpartial{\mathcal{L}}{B} = 0$ in each iteration. Cholesky decomposition is adopted as the matrix in the gradient problem is always positive-definite.

%\begin{eqnarray}
%  \fracpartial{\mathcal{L}^{lc}}{A} &=& HBB^TH^TA - HBM_{ii}^T \\
%  \fracpartial{\mathcal{L}^{lm}}{A} &=& HH^TA - HM_{i0}^T \\
%  \fracpartial{\mathcal{L}^{gb}}{A} &=& M_{0\bar{i}}M_{0\bar{i}}^TA - M_{0\bar{i}}M_{i\bar{i}}^T \\
%  \fracpartial{\mathcal{L}}{A} &=& \fracpartial{\mathcal{L}^{lc}}{A} + \fracpartial{\mathcal{L}^{lm}}{A} + \lambda\fracpartial{\mathcal{L}^{gb}}{A} + \eta I
%\end{eqnarray}

\subsection{Complexity Analysis}
The complexity of the preparation stage is $O(nlogn + k^3)$, while empirically the time expenses are low especially when random partitions are implemented. If $M=I+A$, the average complexity of each section is $O(k \times (deg + iter \times k) \times n_i)$, including both the time in calculating proximity matrices and in optimization.

With separability implemented, SepNE is available to only embed a proportion of nodes. Besides, when small proportions of nodes are requested, the complexity of SepNE in the second stage is irrelevant to the entire scale of the network. This property yields strong scalability of SepNE to super-scale networks.

\section{Experiments}
We evaluated SepNE on several publicly available real-world networks with different sizes and topics, including three document networks and two social networks. Performances over three benchmark tasks were evaluated: (i) matrix reconstruction on document networks, (ii) multi-class classification tasks on document networks and (iii) multi-label classification tasks on social networks.

\subsection{Experiment Setups}

\begin{table} % datasets
  \centering
  \begin{tabular}{l|c|r|r}
    \hline
    Dataset & Directed & Nodes & Links \\
    \hline
    Wiki & directed & 2,405 & 17,981 \\
    Cora & directed & 2,708 &  5,429 \\
    Citeseer & directed & 3,312 & 4,732 \\
    Flickr & undirected & 1,715,255 & 22,613,981 \\
    Youtube & undirected & 1,157,827 & 4,945,382 \\
    Wiki-Gen & directed & $2^i$ & $\approx 7.5\times 2^i$ \\
    \hline
  \end{tabular}
  \caption{Statistics of datasets used in this paper.}\label{tab:data}
\end{table}

\subsubsection{Datasets.} Five real-world networks were used in this paper. \textit{Wiki}, \textit{Cora} and \textit{Citeseer} are thousand-level document networks.
\footnote{Available at \url{https://linqs.soe.ucsc.edu/data}}
Wiki contains Wikipedia pages of $19$ classes; Cora contains machine learning papers from $7$ classes and \textit{Citeseer} contains publications from $6$ classes. Links between documents are pointers or citations. \textit{Flickr} and \textit{Youtube} are million-level social networks.
\footnote{Available at \url{http://socialnetworks.mpi-sws.org/datasets.html}}
Users and their relationships are represented as nodes and links on the networks, and real-world communities are available. \textit{Wiki-Gen}s are a series of networks generated by implementing Kronfit~\shortcite{kronfit} on Wiki. Table~\ref{tab:data} shows the statistics of all datasets.

\subsubsection{Comparison Algorithms and Parameters.} Algorithms and their parameters are briefly introduced below. We did not compare SepNE with algorithms that are not scalable to large networks. Except otherwise noted, the representation dimension for all algorithms was $d=128$.
\begin{itemize}
  \item \textbf{SVD} was conducted over the full proximity matrices. As SVD theoretically generates the optimal rank-$d$ approximations in F-norm, it was proposed as the strongest possible baseline for matrix reconstruction.
  \item \textbf{Nystr\"om Method}~\cite{nystrom} is a fast monte-carlo method of approximating matrix multiplications. It was taken as a representative of probabilistic MF algorithms. The number of \textit{landmarks} in Nystr\"om method is set the same as SepNE for fair comparison.
  \item \textbf{LINE}~\shortcite{line} embeds a network by optimizing an objective function of edge reconstruction. The parameters were set the same as the original paper, namely $\rho_0=0.025$, negative sampling $K=5$ and sample size $T=10^{10}$.
  \item \textbf{DeepWalk}~\shortcite{deepwalk} embeds nodes by regarding them in random walk sequences as words in sentences. The parameters were set as window size $win = 10$, walk length $t = 40$ and walks per node $\gamma = 40$.
  \item \textbf{SepNE} was evaluated under all three partition setups (SepNE-LP, SepNE-RP and SepNE-IO). On document networks, parameters were set as $iter=100$, $\lambda=0.4$ and $\eta=0.1$, $M=A+A^2$ and $k=200$; on social networks, parameters were $iter=5$, $\lambda=50$, $\eta=1$, $M=I+A$ and $k=1000$.
\end{itemize}

\subsection{Running Time}
\begin{table}
  \caption{Running time comparison over flickr.}\label{tab:eff-cmp}
  \centering
  \begin{tabular}{l|r}
    \hline
    SepNE-IO & 6.2mins \\
    SepNE-RP & 43.8mins \\
    SepNE-LP & 68.8mins \\
    LINE(1st) & 138.1mins \\
    DeepWalk & $>$24hrs \\
    \hline
  \end{tabular}
\end{table}
\begin{figure}
  \centering
  \includegraphics[width=0.4\textwidth]{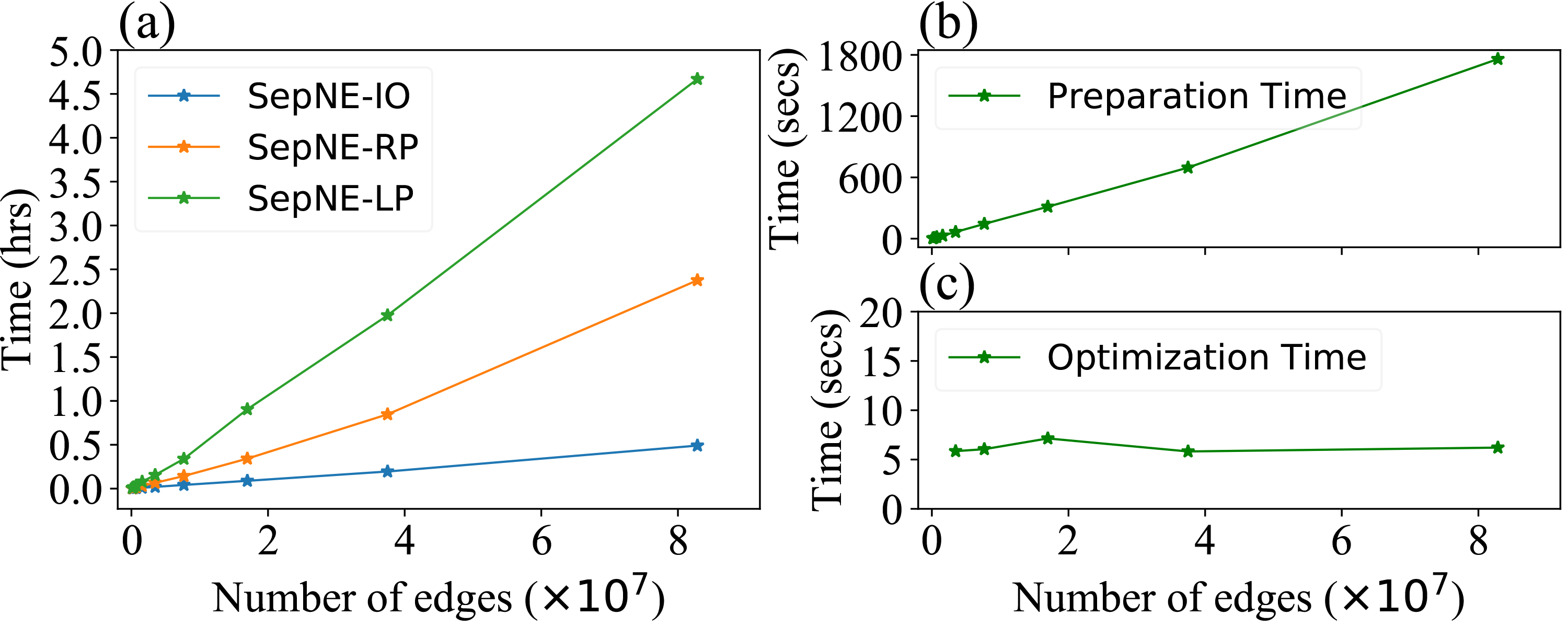}
  \caption{Scalability of SepNE, demonstrated on Wiki-Gens.}\label{fig:eff}
\end{figure}

To demonstrate the speed advantage of SepNE, we compared the running time of SepNE, LINE and DeepWalk on Flickr network. \footnote{All efficiency experiments were conducted on a single machine with $128$GB memory, $32$ cores $2.13$GHz CPU with 16 workers.} The nodes in the five biggest communities of Flickr (75,338 nodes, 4.39\%) were regarded as interested for SepNE-IO. Results are presented in Table~\ref{tab:eff-cmp}. With better performance (introduced below), our method was significantly faster than LINE (for 50.2\%) and DeepWalk even in embedding the entire network; when requested to embed only the nodes of interest, SepNE completed the task in a very short time. \footnote{Data are saved as edge lists in the experiments for fair comparison. If adjacent lists are available, the time of SepNE-IO can still be reduced significantly.}

We also evaluated the trend of running time with network scales increasing on Wiki-Gens and the number of requested nodes for SepNE-IO fixed as 10,000. Figure~\ref{fig:eff} (a) shows a linear trend in all three setups. Figure~\ref{fig:eff} (b)(c) show the trends of preparation and optimization time in SepNE-IO. Preparation time increased linearly mainly due to the time used in reading data, while optimization time remained invariant. The results all corroborate that SepNE is scalable to super-large networks.

\subsection{Matrix Reconstruction}

The performance of reconstructing proximity matrix is a direct metric of representation quality. We evaluated matrix reconstruction performances of different algorithms on all document datasets. As the results were similar, we took Wiki as a representative. Two metrics were used, including the $R^2$ score over all entries ($r_{all}$) and non-zero entries ($r_{nz}$):
\begin{eqnarray*}
  r_{all} &=& 1 - \frac{\Vert \tilde{M} - M \Vert_F^2}{\Vert M \Vert_F^2}, \\
  r_{nz} &=& 1 - \frac{\Vert (\tilde{M} - M) \times B \Vert_F^2}{\Vert M \Vert_F^2}, \\
\end{eqnarray*}
where $\times$ indicates element-wise multiplication and $B_{ij}=1$ if $M_{ij} \ne 0$ otherwise $0$.

\begin{figure}
  \centering
  \includegraphics[width=0.4\textwidth]{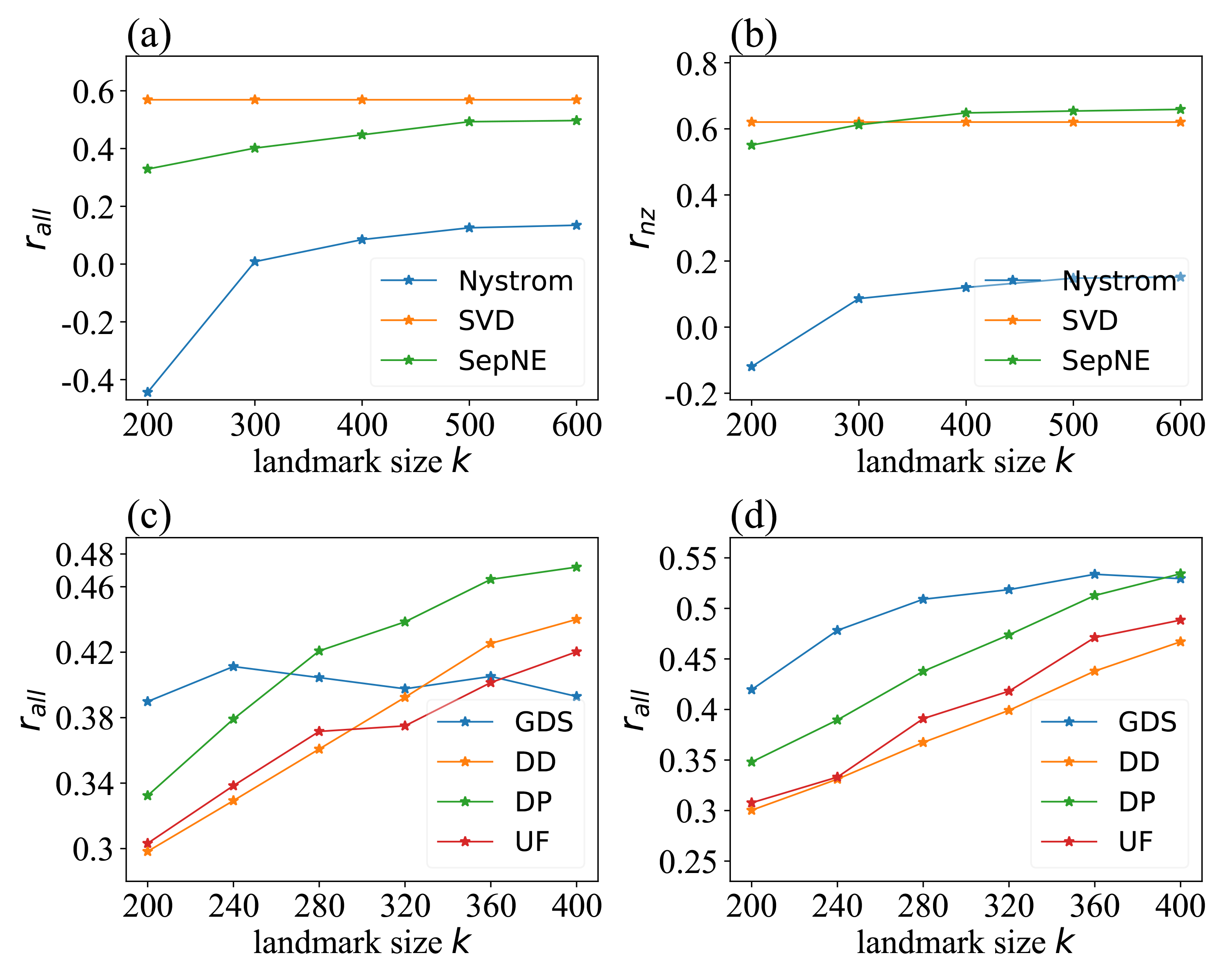}
  \caption{Performances of matrix preserving on Wiki network: (a)(b). $r_{all}$ and $r_{nz}$ of different algorithms versus $k$; (c)(d). $r_{all}$ of different landmark-selecting approaches versus $k$ ($d=128$ in (c) and $200$ in (d)). }\label{fig:mat}
\end{figure}

We evaluated Nystr\"om method, SVD and SepNE with different $k$ over both metrics. We then compared different landmark-selecting approaches with different $k$ and $d$, including four: \textit{DD (Degree Deterministic)} picking nodes with the $k$ highest degrees; \textit{DP (Degree Probabilistic)} sampling landmarks using degrees as weights; \textit{UF (Uniform)} uniformly selecting landmarks; and GDS.

According to Figure~\ref{fig:mat} (a)(b), SepNE significantly outperforms Nystr\"om method for up to $38.3\%$ and shows competent performance compared with SVD. SVD shows its advantage on $r_{all}$, while preserving non-zero entries can be more important than zero entries on real-world networks due to the existence of unobserved links. When $k$ is large enough, SepNE outperforms SVD on $r_{nz}$. This is because non-zero entries are more densely distributed inside communities and better reconstructed by SepNE.

In Figure~\ref{fig:mat} (c)(d), when $k$ is slightly larger than $d$, or to say that $k$ is strictly confined, GDS shows its significant advantages. However, the $r_{all}$ of GDS decreases when $k > 2d$ due to the null representations generated in SVD. At the same time, degree-based approaches gradually get rid of their biases as $k$ increases, and therefore show continuous improvements of performance.

\subsection{Classification}
We implemented two types of classification tasks on document and social networks. A simple logistic regression was used as the classifier for both tasks. The representations were all normalized before used as features. All results were averaged over 10 runs.

\subsubsection{Multi-class classification.} We implemented multi-class classification on three document networks which predicts the subject category a given document is in. Table~\ref{tab:exp-mc} reports the performances. Macro F1 results are not shown due to the similarity. Despite the minimum information SepNE leverages, it outperforms DeepWalk in the majority (4 out of 6) of cases. This is because SepNE incorporates a more robust and elegant way to leverage proximities between nodes. LINE is struggling to capture information on smaller networks, while SepNE is as well competent.

\begin{figure}
  \centering
  \includegraphics[width=0.4\textwidth]{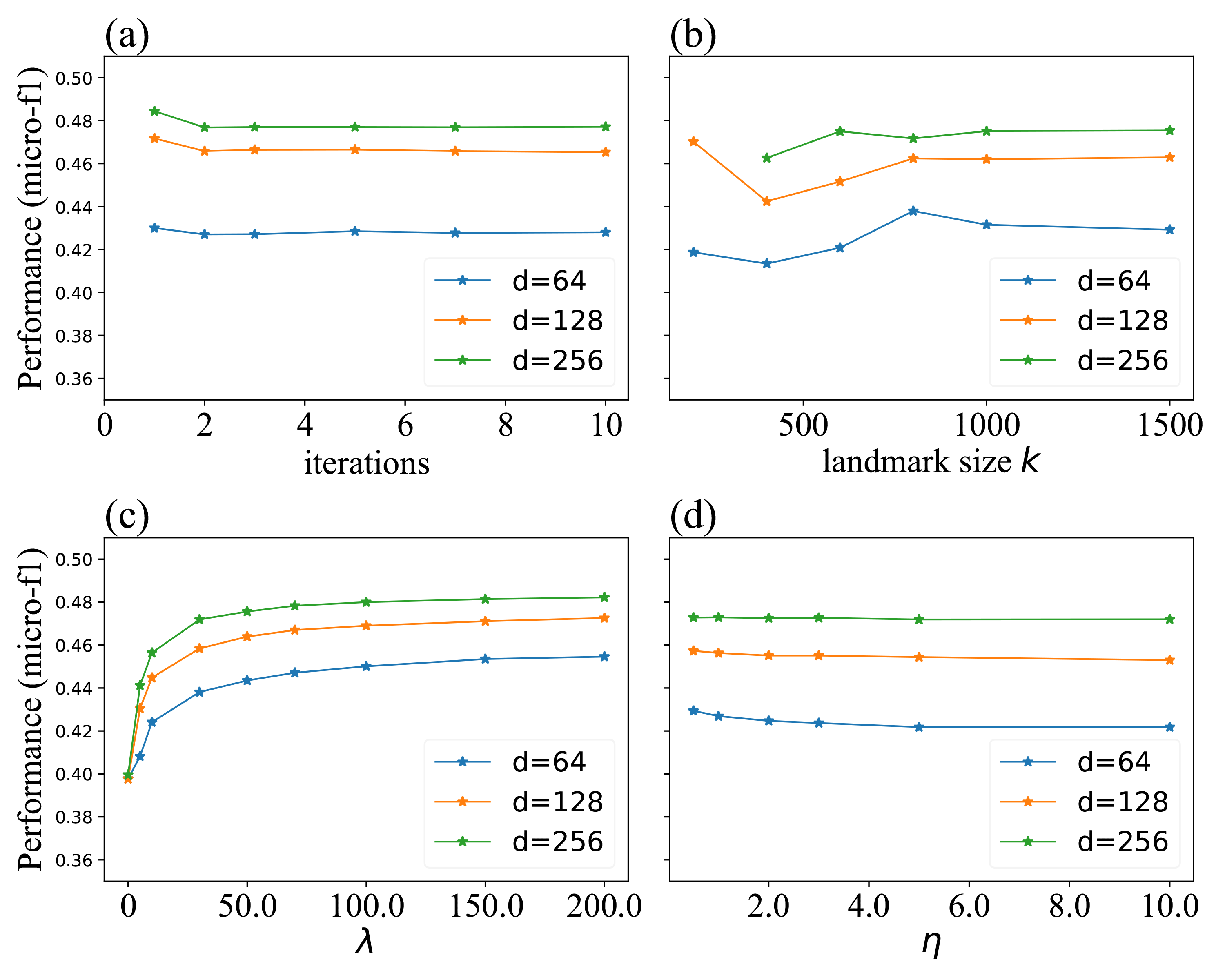}
  \caption{Performances of multi-label classification under different parameter settings on Flickr network.}\label{fig:par}
\end{figure}

\subsubsection{Multi-label classification.} The multi-label classification task on social networks was defined as predicting whether a given node is in each community. The five largest in Flickr and communities with more that 1,000 members in Youtube were extracted as labels. As labels were sparse, we conducted training and predicting processes over the nodes that have at least one label. The training percentage was varied from 1\% to 90\%. Table~\ref{tab:flickr} and Table~\ref{tab:youtube} show the results.

\begin{table*}
  \centering

  %DOCUMENT
  \caption{Multi-class prediction results over document networks \textit{(micro-averaged F1 scores)}. Best performances are bolded.}
  \begin{tabular}{lrrrrrr}
    \hline
    & \multicolumn{2}{c}{Wiki}  & \multicolumn{2}{c}{Cora} & \multicolumn{2}{c}{Citeseer} \\
    \textit{\%train} & 10\% & 90\% & 10\% & 90\% & 10\% & 90\% \\
    \hline
    LINE(1st) & 0.4488  & 0.5937  & 0.4657  & 0.6009  & 0.3206  & 0.4259  \\
    LINE(2nd) & 0.3298  & 0.4787  & 0.2637  & 0.3297  & 0.2221  & 0.2561  \\
    DeepWalk & 0.5737  & \textbf{0.6893}  & \textbf{0.7509}  & 0.8187  & 0.5086  & 0.5813  \\
    SepNE & \textbf{0.5764}  & 0.6867  & 0.7365  & \textbf{0.8220}  & \textbf{0.5157}  & \textbf{0.6072}  \\
    \hline
  \end{tabular}

  %FLICKR
  \label{tab:exp-mc}
  \caption{Multi-label prediction results over Flickr network \textit{(micro-averaged F1 scores)}. Best performances are bolded.}
  \begin{tabular}{lrrrrrrrr}
  \hline
    \textit{\%train} & 1\% & 3\% & 5\% & 10\% & 20\% & 30\% & 50\% & 90\% \\
  \hline
    LINE(1st) & 0.3683 & 0.4118 & 0.4165 & 0.4219 & 0.4270 & 0.4273 & 0.4296 & 0.4274 \\
    LINE(2nd) & 0.3450 & 0.3824 & 0.3955 & 0.3973 & 0.4032 & 0.4056 & 0.4069 & 0.4068 \\
    DeepWalk & 0.4072 & 0.4353 & 0.4433 & 0.4481 & 0.4518 & 0.4564 & 0.4585 & 0.4592 \\
    SepNE-IO & 0.4065 & 0.4341 & 0.4477 & 0.4562 & 0.4582 & 0.4607 & 0.4630 & 0.4622 \\
    SepNE-RP & 0.4061 & 0.4388 & 0.4502 & 0.4601 & 0.4628 & 0.4634 & 0.4636 & 0.4658 \\
    SepNE-LP & \textbf{0.4269} & \textbf{0.4468} & \textbf{0.4562} & \textbf{0.4623} & \textbf{0.4645} & \textbf{0.4656} & \textbf{0.4674} & \textbf{0.4677} \\
  \hline
  \end{tabular}
  \label{tab:flickr}

  %YOUTUBE
  \caption{Multi-label prediction results over Youtube network \textit{(micro-averaged F1 scores)}. Best two performances are bolded.}
  \begin{tabular}{lrrrrrrrr}
  \hline
    \textit{\%train} & 1\% & 3\% & 5\% & 10\% & 20\% & 30\% & 50\% & 90\% \\
  \hline
    LINE(1st) & 0.1031 & 0.2322 & 0.2745 & 0.3141 & 0.3410 & 0.3520 & 0.3594 & 0.3673 \\
    LINE(2nd) & 0.0782 & 0.1839 & 0.2158 & 0.2643 & 0.2987 & 0.3159 & 0.3280 & 0.3350 \\
    DeepWalk & 0.2037 & \textbf{0.3397} & \textbf{0.3739} & \textbf{0.4105} & \textbf{0.4355} & \textbf{0.4438} & \textbf{0.4501} & \textbf{0.4556} \\
    SepNE-IO & 0.2035 & 0.3325 & 0.3574 & 0.3885 & 0.4041 & 0.4129 & 0.4170 & 0.4216 \\
    SepNE-RP & \textbf{0.2256} & 0.3355 & \textbf{0.3633} & \textbf{0.3920} & 0.4115 & 0.4157 & 0.4214 & 0.4273 \\
    SepNE-LP & \textbf{0.2253} & \textbf{0.3361} & 0.3620 & 0.3882 & \textbf{0.4118} & \textbf{0.4170} & \textbf{0.4218} & \textbf{0.4277} \\
  \hline
  \end{tabular}
  \label{tab:youtube}
\end{table*}

SepNE shows significant advantages on Flickr. Using 10\% training data, SepNE-LP outperforms LINE and DeepWalk using 90\%. Representations from SepNE are more predictive than DeepWalk even if only one-hop proximity is leveraged. The reason may be that as Flickr has relatively high average degree, the larger window size of DeepWalk actually encumbers it in determining the importance of information. All three setups of SepNE show good performances, while SepNE-LP shows its advantage over the other two setups. This shows the effectiveness of the empirical method of partitioning networks according to communities, while the time cost of SepNE-LP is significantly higher than the other two simplified setups. The task on Youtube is more challenging as both the network and labels are much sparser. DeepWalk outperforms both LINE and SepNE due to its ability in leveraging remote proximities with its larger window size, which successfully relieves the problem of sparsity at the cost of much higher time expenses. SepNE outperforms both LINE(1st) and LINE(2nd), which again corroborates its stronger ability to leverage near proximities.

\subsection{Parameter Sensitivity}

Figure~\ref{fig:par} shows the effect of $iter$, $k$, $\lambda$, $\eta$ and $d$. $Iter$, $k$ and $\eta$ do not show significant influences. The performance of SepNE is good with even one iteration, probably indicating that local information on Flickr is less important. Figure~\ref{fig:par} (c) shows that larger $\lambda$ generally leads to better performance, converging with $\lambda \ge 100$. The higher performances of larger $\lambda$s, particularly compared with $\lambda = 0$, show the effectiveness of the elaborated global loss.

\section{Related Work}

There are massive literature proposed over NE problems. Traditional dimension reduction approaches \cite{lle,isomap,laplace} are applicable on network data through Graph Laplacian Eigenmaps or proximity MF. Recently, various skip-gram-based NE models and applications were proposed \cite{deepwalk,node2vec,dne}. Besides, the pioneering work of \cite{wordemb-mf} proved an equivalency between skip-gram models and matrix factorization, which further leads to new proximity metrics under the proximity MF framework \cite{tadw,grarep,netmf}. Edge reconstruction algorithms \cite{line} were proposed to gain scalability on large networks. Neural networks, including autoencoders \cite{sdne,cui4} and CNNs \cite{gcn,graphsage} were also leveraged in NE problems. There is also a new trend \cite{struct2vec} that leverages structural information instead of proximity in NE.

The most similar work to ours is \cite{gf}, in which a similar partition was adopted to achieve separability, while other parts of the work had major differences with ours. Besides, it focused mainly on technical issues in distributed learning and preserved only link information, while SepNE is more generalized idea with a more elaborated optimization goal.

\section{Conclusion}

In this paper, we formalized the problem of separated matrix factorization, based on which we proposed SepNE, an separable network embedding method which outperforms strong baselines in both efficiency and performance.

The key contribution of SepNE is providing a novel perspective of evaluating network embedding methods: separability. A separable method is stronger than a distributable one, as partly conducting a separable task provides meaningful outputs. This property provides an option of embedding only a proportion of nodes and yields strong significance in distributed learning, super-large network embedding and dynamic network embedding. Furthermore, SMF reduces the complexity of MF from cubic to linear with the generalizability over all MF-based algorithms.

SepNE is still a simple framework. For future work, one intriguing direction is to incorporate more complex information into the framework without loss of efficiency. Also, a theoretical proof of a lower bound over the loss in matrix reconstruction, and a more theoretical explanation of SMF can be extremely informative. Should there be such work, it will be theoretically-founded to apply SMF on all matrix-factorization-based algorithms.

%References and End of Paper
%These lines must be placed at the end of your paper
\nocite{largedata}
\nocite{cui1}
\nocite{cui2}
\nocite{cui3}
\bibliography{ref}
\bibliographystyle{aaai}
\end{document}